\documentclass{workshop2008}

\usepackage{graphicx}

\newcommand{\RPR}{\textit{R\underline{P}R}}
\newcommand{\PRR}{\textit{\underline{P}RR}}

\begin{document}

\title{Changing Assembly Modes without Passing Parallel Singularities in Non-Cuspidal 3-R\underline{P}R Planar Parallel Robots}

\author{Ilian A.~Bonev, S\'ebastien Briot
\affiliation{D\'epartement de g\'enie de la production automatis\'ee\\
\'Ecole de technologie sup\'erieure\\
1100, rue Notre-Dame Ouest\\
Montreal, QC, Canada H3C 1K3}}

\author{Philippe Wenger, Damien Chablat
\affiliation{Institut de Recherche en Communications et\\
 Cybern\'etique de Nantes UMR CNRS 6597\\
1, rue de la No\"e, BP 92101\\
44312 Nantes Cedex 03 France}}

\maketitle

\begin{abstract}
This paper demonstrates that any general 3-DOF three-legged planar parallel robot with extensible legs can change assembly modes without passing through parallel singularities (configurations where the mobile platform loses its stiffness). While the results are purely theoretical, this paper questions the very definition of parallel singularities.
\end{abstract}

\section{Introduction}

Most parallel robots have singularities that limit the motion of the mobile platform. The most dangerous ones are the singularities associated with the loss of stiffness of the mobile platform, which we call here \textit{parallel singularities}. Indeed, approaching a parallel singularity also results in large actuator torques or forces. Hence, these singularities should be avoided for most applications. A safe solution is to eliminate parallel singularities at the very design stage [1,2] or to define singularity-free zones in the workspace [3,4]. It is also possible to avoid parallel singularities when planning trajectories [5,6].

For a parallel robot with multiple inverse kinematics solutions, belonging to different \textit{working modes}, a change of configuration of one of its legs may allow it to avoid a parallel singularity [7,8]. This paper addresses a recent, difficult issue that has been investigated by few researchers: the possibility for a parallel robot to move between two direct kinematic solutions, belonging to two \textit{assembly modes}, without encountering a parallel singularity. We will focus on planar parallel robots with three extensible legs, referred to 3-\RPR\footnote{\textit{R} and \textit{P} stand for revolute and prismatic joints, respectively. The underlined letter refers to the actuated joint.}. As shown in [9], the study of the 3-\RPR\ planar robot may help better understand the kinematic behavior of its more complex spatial counterpart, the octahedral hexapod.

Planar parallel robots may have up to six direct kinematic solutions (or assembly-modes). It was first pointed out that to change its assembly-mode, a 3-\RPR\ planar parallel manipulator should cross a parallel singularity [10]. But [11] showed, using numerical experiments, that this statement is not true in general. In fact, an analogous phenomenon exists in serial robots, which can move from one inverse kinematic solution to another without meeting a singularity [11]. The non-singular change of posture in serial robots was shown to be associated with the existence of points in the workspace where three inverse kinematic solutions meet, called cusp points [12]. On the other hand, McAree and Daniel [9] pointed out that a 3-\RPR\ planar parallel robot can execute a non-singular change of assembly-mode if a point with triple direct kinematic solutions exists in its joint space. The authors established a condition for three direct kinematic solutions to coincide and showed that a non-singular assembly-mode changing trajectory in the joint space should encircle a cusp point.

Wenger and Chablat [13] investigated the question of whether a change of assembly-mode must occur or not when moving between two prescribed poses in the workspace. More recently, Zein et al. [14] investigated the non-singular change of assembly-mode in planar 3-\RPR\ parallel robots and proposed an explanatory approach to plan non-singular assembly-mode changing trajectories by encircling a cusp point. Finally, the most recent results showed that a non-singular change of assembly-mode is possible without moving around a cusp point [15,16].

In [15], an example of a 3-\RPR\ planar parallel robot was given whose workspace is divided into two portions by the singularity surface, while having more than two assembly modes. It is therefore obvious that the robot can change an assembly mode to at least another one without crossing any singularity. In [16], an example of a 3-\PRR\ planar parallel robot (with actuators having parallel directions) was given and it was shown that an assembly mode can be changed by passing through a serial singularity (in which a leg is singular) and changing working modes.

In this paper, we show that any non-architecturally singular 3-\RPR\ planar parallel robot can change assembly-mode without encircling a cusp point or passing through a parallel singularity.

\section{Singularity Loci of 3-\RPR\ Planar Parallel Robots}

Referring to Fig.~1, we denote with $A_i$ and $B_i$ the base and platform revolute joints, respectively. The directed distance between $A_i$ and $B_i$ along the direction of prismatic actuator $i$ is $\rho_i$, which is the active joint variable. Finally, we denote by $C$ the center of the mobile platform.

\begin{figure}
	\centering
		\scalebox{1}{\includegraphics{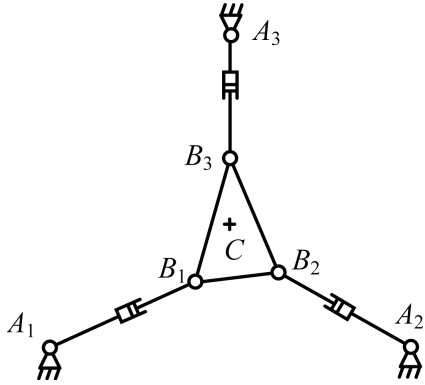}}
	\label{fig:1}
	\caption{Schematics of a general 3-\RPR\ planar parallel robot}
\end{figure}

It is well known that the 3-\RPR\ planar parallel robot is at a parallel singularity when the lines passing through the passive revolute joints in each leg intersect at one point or are parallel. These lines represent the reciprocal screws, i.e., the reaction forces applied to the mobile platform [17].

The singularity loci of this robot, defined as the set of positions of point $C$ where the robot is at singularity for a given orientation, were studied in detail in [18] and it was shown that they form a conic (i.e., a hyperbola, a parabola or an ellipse), unless there is an architectural singularity. An architectural singularity occurs, for example, when the mobile platform and the base form similar triangles, in which case there is an orientation at which all positions correspond to singularities. A more general study of the singularity surface of a 3-\RPR\ parallel robot is given in [19].

In [17], it was shown that all points from this conic correspond to parallel singularities except for three (or two, or one) of them. These three points correspond to the poses of the platform in which one (or more) legs are in a serial singularity, i.e., in which two revolute joints in a leg coincide. Such a singularity corresponds to an uncontrollable passive motion [18]. In such a configuration, the reciprocal screw associated with the singular leg degenerates to two linearly independent forces. Thus, in such a configuration, there is a parallel singularity if and only if the lines associated with the two non-singular legs pass through the coinciding revolute joints of the singular leg. This would only be possible for special designs in which an angle of the base triangle is equal to the corresponding angle of the platform triangle. 

That passing the singularity curve means passing though a parallel singularity, as shown in Fig.~2, is fairly common knowledge. What no one has previously pointed out is that this singularity curve has passages that usually correspond to serial singularities only. Figure~3 illustrates that virtually any general 3-\RPR\ planar parallel robot can cross such a singularity curve through these special passages, without even being near a parallel singularity (though measuring proximity to a parallel singularity is still an open question).

\begin{figure*}[T]
	\centering
		\scalebox{1}{\includegraphics{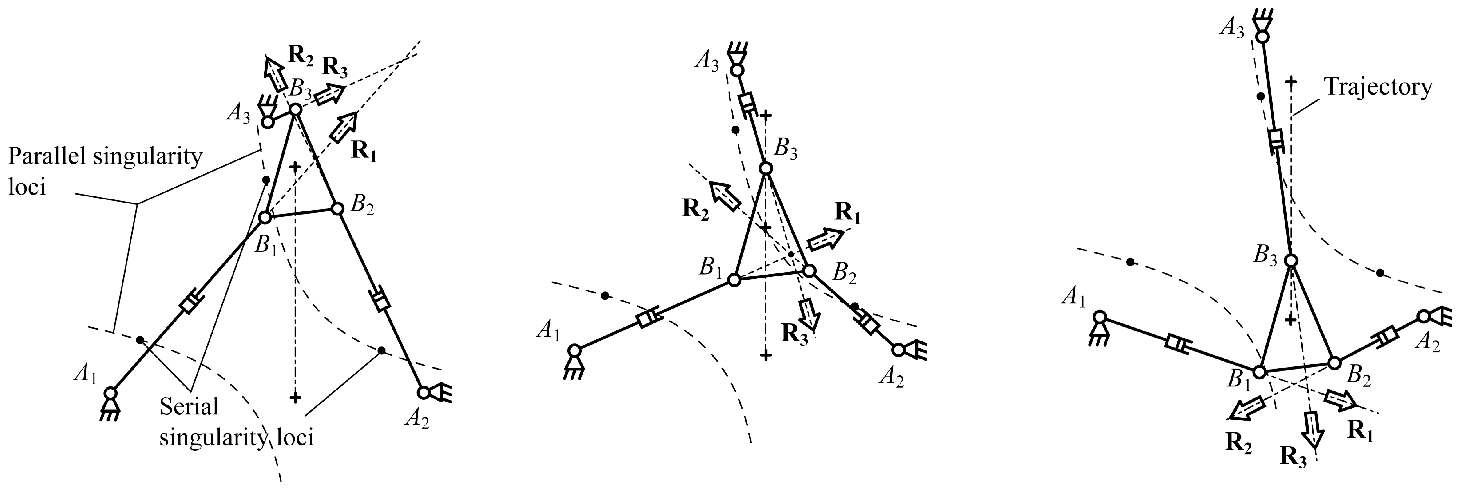}}
	\label{fig:2}
	\caption{Crossing the singularity curve generally means passing through a parallel singularity}
\end{figure*}

\begin{figure*}[t]
	\centering
		\scalebox{1}{\includegraphics{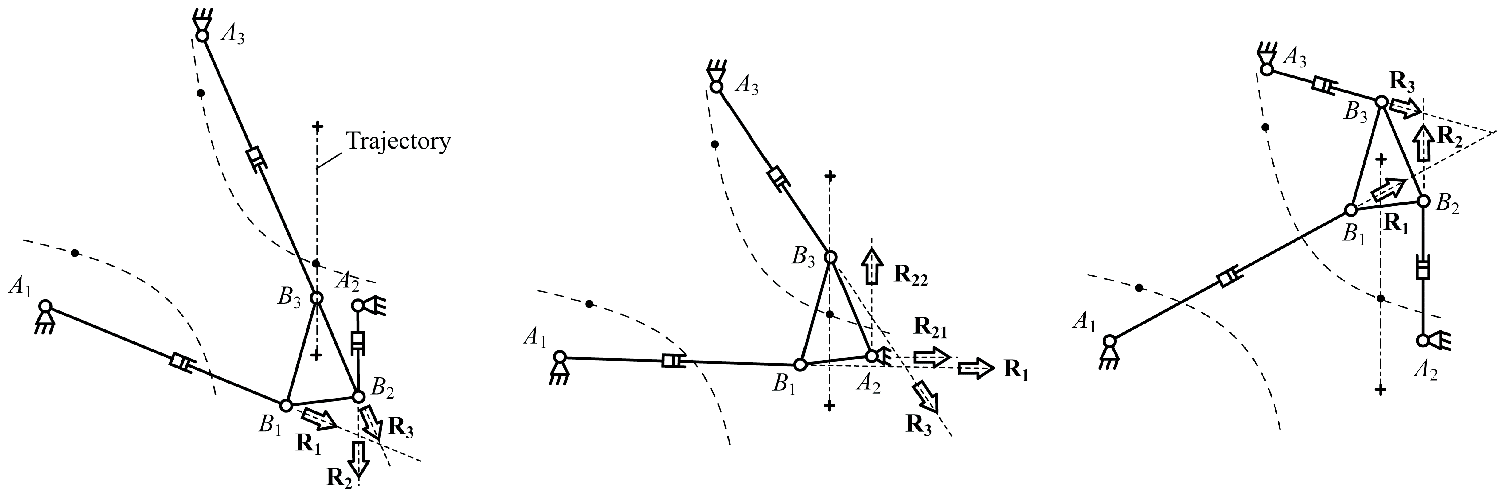}}
	\label{fig:3}
	\caption{Crossing the singularity curve without passing through a parallel singularity}
\end{figure*}

Since this robot can cross the singularity curves for any orientation without passing through a parallel singularity, it is obvious that it can switch between any two assembly modes, without passing through a parallel singularity.

\section{What Exactly is a Parallel Singularity?}

Of course, the main purpose of this paper is purely theoretical. It is clear that, in practice, a 3-\RPR\ parallel robot could hardly pass a serial singularity for many reasons. The most obvious one seems to be that a prismatic actuator that can change from positive lengths to negative ones can be difficult to build. However, this problem can be easily overcome if we use \textit{R\underline{R}R} legs instead, so this is not an issue. What is more difficult is to cope with the (uncontrollable) passive motion of the singular leg. If such a passive motion occurs during a serial singularity, then the prescribed trajectory of Fig.~2 can no longer be followed. Finally, it would be even more difficult to drive the robot to such a configuration. Outside serial singularities, the 3-\RPR\ parallel robot can accept input errors --- this would simply result in output errors. However, in a serial singularity, no errors are possible --- this would result in jamming (the lengths of the two non-singular legs are not independent from one another). Note that this is not the case in most other parallel robots: they can cross a serial singularity without any difficulty.

This brings us to the essential question of what is a parallel singularity. The mobile platform clearly does not lose stiffness in the configuration in question. Yet, in this configuration, the number of direct kinematic solutions abruptly drops to one (or two for some special cases). Indeed, the direct kinematic problem of this robot comes to finding the maximum six intersection points between a sextic and a circle. When an actuator has zero length, this circle degenerates to a point, which explains the sudden drop in the number of solutions: there may be at most two ``intersection points'' between a point and a sextic.

This interpretation also helps understand why such a configuration is not tolerant to input errors. In the case of non-zero leg lengths, if we slightly change the lengths of the two legs corresponding to the sextic, the latter will slightly change but there could still be six intersection points. If however, we do the same in the case of a zero-length leg, the point that is the degeneration of a circle, will no longer lie on the sextic.

\section{Conclusions}

This paper demonstrates that the problem of assembly-mode changing is still an open issue and its objectives should be better defined. Namely, the restrictions on such a change should be specified. Can we pass a serial singularity? If the answer is positive, should we be able to do this in practice or not necessarily?

This paper also questions the very definition of a parallel singularity, often associated with both a loss of stiffness and degeneracy of the direct kinematics. An example of a serial singularity is given in which the direct kinematics degenerates but the mobile platform does not lose stiffness.

\end{document}